\RequirePackage{fix-cm}

\documentclass[twocolumn]{svjour3}          

\smartqed  

\usepackage{tikz}
\usepackage{graphicx}
\usepackage{hyperref}
\usepackage{authblk}
\usepackage[numbers]{natbib}
\usepackage{amsmath}
\usepackage{amssymb}
\usepackage{bbm}
\usepackage{commath}
\usepackage{algorithmic}
\usepackage[linesnumbered, ruled, noline, noend]{algorithm2e}
\usepackage{tabularx}
\usepackage{booktabs}
\usepackage{textcomp}
\usepackage{enumitem}
\usepackage{latexsym}

\usepackage{longtable}
\usepackage{mathrsfs}

\usepackage{orcidlink}
\usepackage{hyperref}

\newcolumntype{L}{>{$}l<{$}}
\newcolumntype{C}{>{$}c<{$}}
\newcolumntype{R}{>{$}r<{$}}

\newcommand*{\affaddr}[1]{#1}
\newcommand*{\affmark}[1][*]{\textsuperscript{#1}}

\usepackage{makecell, caption, booktabs, multirow, siunitx}

\newcommand{\otoprule}{\midrule[\heavyrulewidth]}
\usepackage{lipsum, fancyvrb}
\usepackage{bbm}
\usepackage{marvosym}
\usepackage{tikz}
\usepackage{xcolor}
\begin{document}

\title{Rethinking Fisher Vectors for Second-order Deep Feature Aggregation in Scalable Image Retrieval}

\titlerunning{Rethinking Fisher Vectors for Second-order Deep Feature Aggregation in Scalable Image Retrieval}     

\author{
Alexy Bhowmick \orcidlink{0000-0003-2334-5856} \Letter {\protect\affmark[1]} \and Sarat Saharia \orcidlink{0000-0001-8026-3174} \affmark[2] \and Shyamanta M. Hazarika \orcidlink{0000-0003-4547-6013}\affmark[3]}

\authorrunning{A. Bhowmick et al.} 

\institute{\affmark[1]Department of Computer Science and Engineering, Assam Don Bosco University, Azara, Guwahati 781017, Assam, India. 
              \email{alexy.bhowmick@gmail.com}  
           \and \\
\affaddr{\affmark[2]Department of Computer Science and Engineering, Tezpur University, Tezpur 784028, Assam, India}\\
\affaddr{\affmark[3]Biomimetic Robotics and Artificial Intelligence Lab, \\ Department of Mechanical Engineering, Indian Institute of Technology, Guwahati 781039, Assam, India}\\
}

\date{Received: date / Accepted: date}

\maketitle

\begin{abstract}
Aggregating CNN responses into a compact discriminative vector, given their lack of geometric invariance, is a challenge. To address this challenge, we propose a visual vocabulary-free method to encode second-order information of multi-scale CNN features within the Fisher Kernel framework. We propose a scalable retrieval pipeline to show the efficacy of the proposed vector. The increased discriminative power is confirmed by experimental evaluation of our method on two difficult benchmarks (INRIA Holidays, UKBench). Systematic experiments and ablation studies with multiple backbones demonstrate the proposed method achieving a mAP of 83.42\% on INRIA Holidays and a N-S score of 3.76 on the UKBench datasets respectively. The significant improvement in retrieval accuracy over state-of-the-art approaches comes with only a slight increase in dimension. The proposed signature vector can also be expected to improve image and object classification performance.

\end{abstract}

\section{Introduction}
\label{intro}
Scene retrieval has witnessed a steady stream of advancements and has drawn a sustained attention from the computer vision and multimedia field for decades. The goal of scene retrieval is to retrieve from the dataset images that depict the same scene or objects. Most of the conventional scene retrieval algorithms encode low-level, hand-crafted features of an image or video into a comprehensive, short vector representation. It has prompted the development of several notable encoding schemes, including the Vector of Locally Aggregated Descriptors (VLAD) \cite{jegou_aggregating_2012}, Fisher Vector (FV) \cite{perronnin_fisher_2007}, and Triangulation Embedding (Temb) \cite{jegou_triangulation_2014}. These techniques successfully achieve an accuracy-efficiency trade-off but fall short in terms of capturing high-level semantic data for scene retrieval. Therefore, it is essential to develop efficient scene retrieval solutions that include the combined optimization of the fundamental constraints: \textit{accuracy}, \textit{speed}, and \textit{memory footprint}.

\begin{figure*}[htp]
	\centering
	\includegraphics[width=\textwidth]{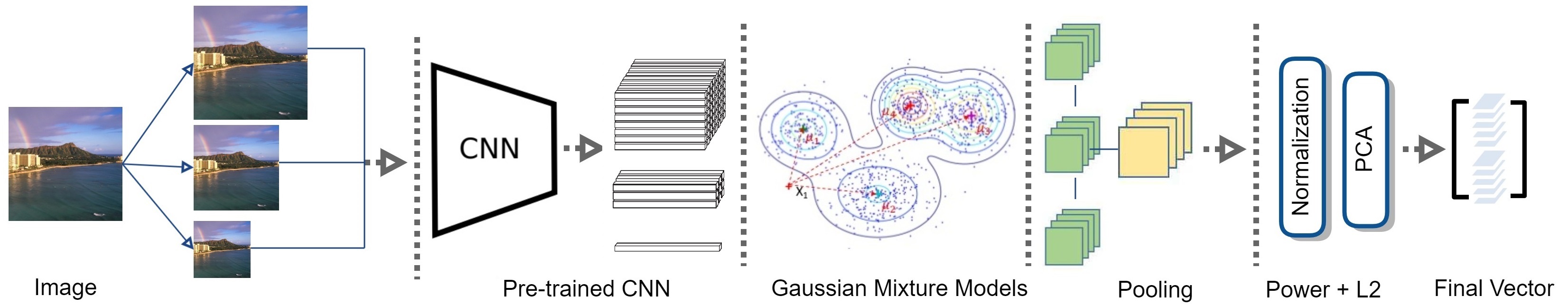}
	\caption[Overview of the proposed codebook-free scheme for second-order multi-scale encoding and pooling of CNN-activations.]{Overview of the proposed codebook-free scheme for second-order multi-scale encoding and pooling of CNN-activations.}
	\label{fv-cnn-encoding}
\end{figure*}

Recently, the focus of research has shifted towards CNN image representations which have become state-of-the-art generic descriptors for image classification and retrieval. As they capture high-level semantic knowledge of objects and surroundings, convolutional features offer greater discriminative power than traditional hand-crafted features. Convolutional features also handle the effect of \textit{burstiness} which is known to occur in the local features. The higher discriminative power enables CNNs to exhibit superior performance and significantly reduce the \textit{semantic gap}. Recent retrieval techniques \citep{babenko_neural_2014, gong_multi-scale_2014, razavian_cnn_2014, bhowmick_fhvlad_2021, wang_aggregating_2023} have demonstrated that activations from the intermediate layers of a pre-trained CNN work well as high-level image descriptors. In one of the early studies, Razavian et al. \citep{razavian_cnn_2014} found that CNN features without any embedding or complex aggregation were less effective. Consequently, the authors incorporated CNN features with spatial information to achieve competitive performance in image retrieval. 

For tasks like visual recognition \cite{chatfield_return_2014}, semantic segmentation \cite{badrinarayanan_segnet_2017}, etc., representations based on the activations of layers of such networks have proven successful. There is evidence that deeper layer activations produce superior image representations \citep{uricchio_fisher_2015}; but it is still being explored how to aggregate these responses into a fixed-length discriminative vector for image retrieval and classification. While a majority of techniques simply extract activations from the network's penultimate layers, another line of research has looked into aggregating higher-order statistics of CNN activations for scene retrieval. However, global CNN activations do not exhibit geometric variance \citep{gong_multi-scale_2014}, which reduces their effectiveness in matching different scenes. Given that Fisher Vectors are an advanced encoding scheme, a number of recent works (e.g.\citep{simonyan_deep_2014}, \citep{uricchio_fisher_2015}, have investigated aggregation of deep features for a robust representation for vision tasks.

Our research focuses on encoding CNN feature maps in order to create a compact image supervector for scalable scene retrieval. We propose a novel method to improve the invariance of CNN features without degrading their discriminative power. We encode the CNN activations from multiple image scales in the Fisher kernel framework \citep{perronnin_large-scale_2010}. In memory-constrained settings, it is preferred to do away with the codebook (visual vocabulary), which is made possible by the Gaussian Mixture Model encodings. The Fisher kernel-inspired second-order aggregation of CNN activations obtained from multiple scales provides notable improvements over the original (baseline) Fisher Vector as well as recent state-of-the-art methods. The discriminative power comes from the exploitation of CNN features and Fisher Vector's rudimentary vector operations which makes it less expensive to compute. Our experiments help to understand how the convolutional features extracted from well-known networks (variants of VGG \cite{simonyan_very_2015} and ResNet \cite{he_deep_2016}) can be aggregated in a sophisticated manner to achieve a significant boost in retrieval performance. 




The main contributions in this paper are: 
\begin{enumerate}
    
    \item A novel visual vocabulary-free image representation that aggregates multi-scale second-order CNN features within the Fisher framework resulting in greater discriminative power. 

    \item A scalable image retrieval framework based on fast indexing and effective search to produce ranked retrieval results.
    
    \item Experiments on two benchmark datasets show that the proposed representation significantly improves retrieval accuracy. 
\end{enumerate} 

The remainder of the paper is structured as follows. Section \ref{sec2} briefly outlines the background and relevant work. In Section \ref{sec3} we describe the encoding algorithm and introduce the novel representation vector. Section \ref{sec4} presents the evaluation process, datasets, ablation study, experimental comparison, and analyses. In Section \ref{sec5},  we examine the limitations of our approach and outline directions for future work. Finally, Section \ref{sec6} concludes the paper. 

\begin{figure*}[htp]
	\centering
	\includegraphics[width=0.9\textwidth]{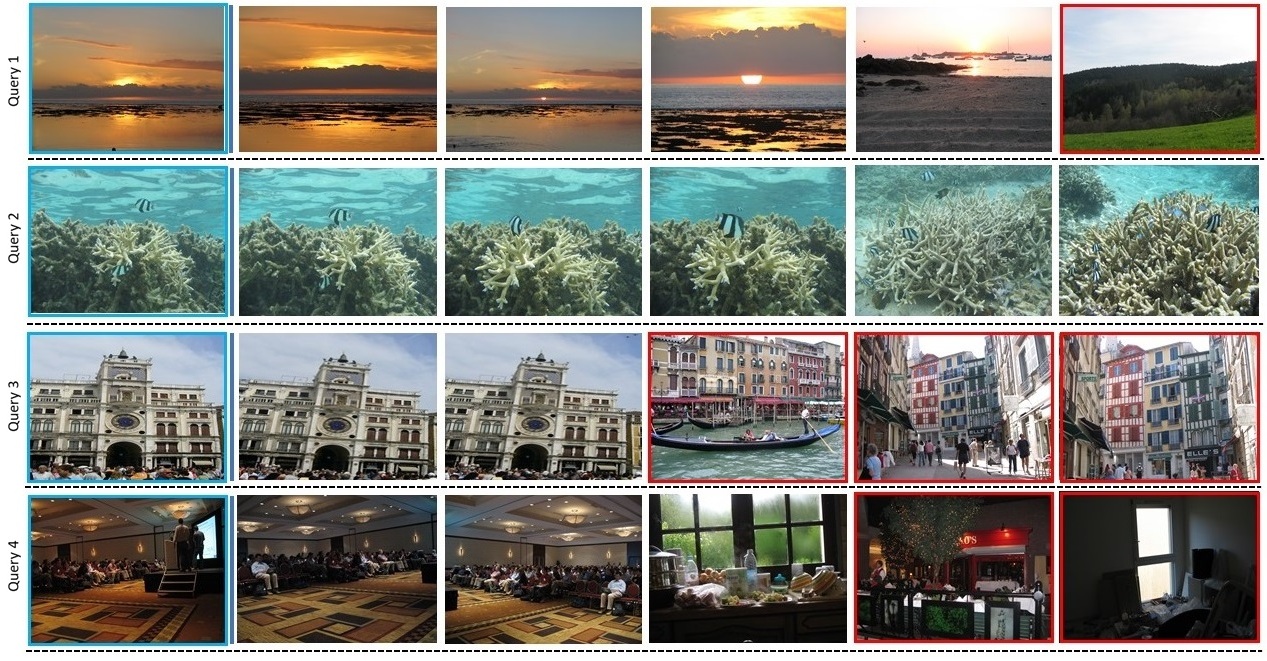}
	\caption[Example baseline Fisher Vector (FV) results on INRIA Holidays \citep{jegou_improving_2010}.]{Accurate retrieval results by baseline Fisher Vector (FV) on INRIA Holidays. Results that do not reflect groundtruth (or are edge cases) are in a red box \fcolorbox{red}{white}{\phantom{\rule{5pt}{5pt}}}. A blue box \fcolorbox{teal}{white}{\phantom{\rule{5pt}{5pt}}}  surrounds each query.}
	\label{fv-accurate}
\end{figure*}
\begin{figure*}[htp]
	\centering
	\includegraphics[width=0.9\textwidth]{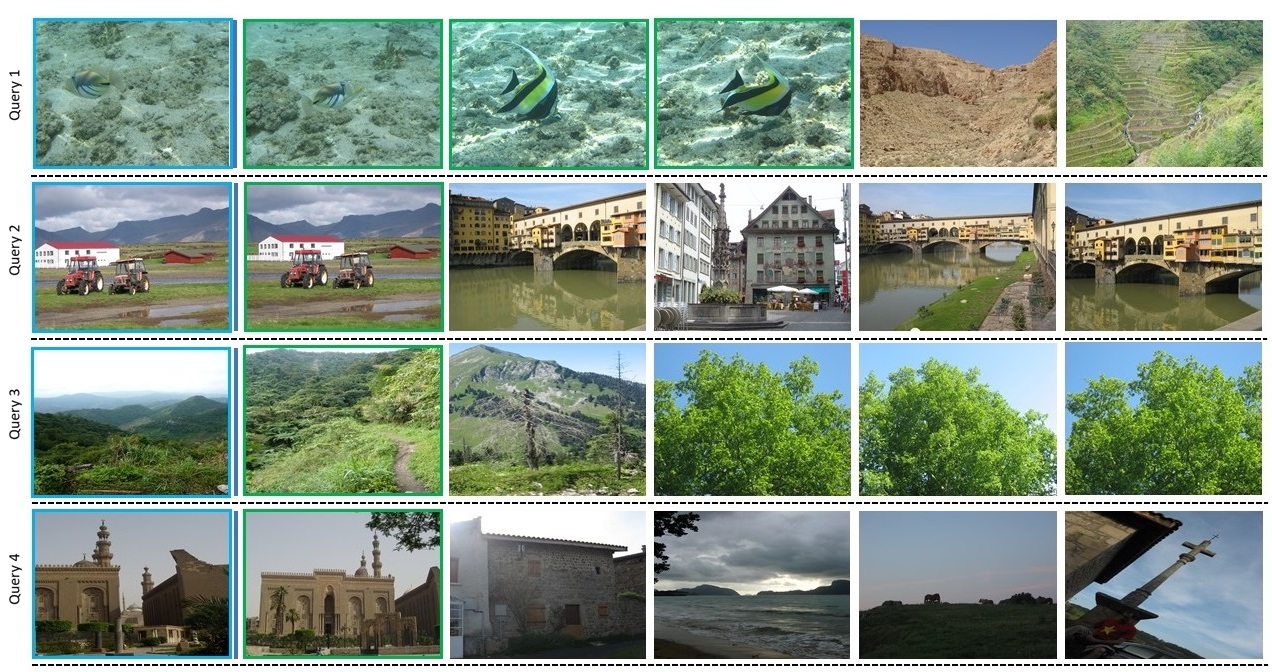}
	\caption[Example baseline Fisher Vector (FV) results on INRIA Holidays \citep{jegou_improving_2010}.]{Sub-optimal retrieval results by baseline Fisher Vector (FV) on INRIA Holidays. Results that reflect groundtruth are in a green box \fcolorbox{green}{white}{\phantom{\rule{5pt}{5pt}}}. A blue box \fcolorbox{teal}{white}{\phantom{\rule{5pt}{5pt}}} surrounds each query.}
	\label{fv-suboptimal}
\end{figure*}
\section{Related Work}
\label{sec2}
This section examines the background and state-of-the-art deep feature vector aggregation methods for scalable image retrieval.

\textbf{CNN Descriptors for Scene Retrieval:} In a seminal study on the use of CNN features for scene retrieval, Wan et al. \citep{wan_deep_2014} employed CNN features on different scene retrieval datasets with model retraining and similarity learning. The findings showed that CNN features are indeed effective in reducing the \textit{semantic gap} that exists between low-level visual features and high-level semantics. The final layer of the CNN was believed to be crucial for strong performance in retrieval. In keeping with this line of inquiry, Babenko et al. \citep{babenko_neural_2014} extensively evaluated the performance of CNN descriptors for scene retrieval and reported that Principal Component Analysis (PCA)-compressed deep features outperformed compact features based on SIFT \cite{lowe_distinctive_2004}. The authors retrained a CNN on a landmark database for their scene retrieval experiments. Unfortunately, the training time of the model made the approach computationally infeasible for practical use. Recently, Bhowmick at al. \cite{bhowmick_encoding_2019} proposed a high-order discriminative encoding of VLAD features which captured complementary high-order descriptor statistics to enhance feature discriminability.

Contrary to earlier assumptions, Razavian et al. \cite{razavian_cnn_2014} found that the traditional SIFT features encoded in VLAD encoding \cite{jegou_aggregating_2010} were superior to CNN features derived from the final layer, and needed to be boosted with spatial features for obtaining comparable results. In a series of experiments, the authors investigated the use of off-the-shelf CNN features and reported good results in visual instance retrieval without aggregating them into a global descriptor. The authors used the L2-normalized output of the first fully connected layer of the fine-tuned \textit{Overfeat} CNN network as a generic representation for the image. The \textit{Overfeat} network was trained for the ImageNet Large Scale Visual Recognition Challenge (ILSVRC 2013) challenge to perform object classification. Although features from the final layers proved less crucial than initially assumed, the results confirmed the effectiveness of CNN representation for visual object recognition and retrieval. 

\textbf{Encoding CNN Descriptors as Compact Vectors:} 
The \textit{Multi-scale Orderless Pooling} (MOP-CNN) \citep{gong_multi-scale_2014} is a significant method that was proposed to aggregate multi-scale CNN activations from the higher layers (Fully Connected layer) and achieve geometric invariance. The aggregation of the features in a VLAD-like \citep{jegou_aggregating_2012} fashion on a multi-scale setting showed promising results. However, it did not address the question of which scales and layers result in the optimum performance for retrieval tasks. Inspired by MOP-CNN's \citep{gong_multi-scale_2014} VLAD-like aggregation of CNN features, Ng et al. \citep{ng_exploiting_2015} extracted convolutional features from network's layers (at different scales) for instance-level scene retrieval, and demonstrated that the intermediate layers with finer scales performed better than final layers in case of instance-level scene retrieval. The authors used VLAD encoding \citep{jegou_aggregating_2012} to encode low-dimensional representations of CNN responses. Notably, the new feature preserved local object characteristics, captured category-level information, and achieved state-of-the-art retrieval outcomes. 

In a systematic study, Azizpour et al. \cite{azizpour_generic_2015} looked into various factors related to CNN representations and their transferability including parameters and architecture. The \textit{sum-pooled convolutional features} (SPoC) descriptor \citep{babenko_aggregating_2015}, based on raw deep convolutional features, from the very deep trained ConvNet configuration by Simonyan et al. \citep{simonyan_very_2015}, is a significant pooling strategy. SPoC introduced sum-pooling of deep features instead of max-pooling. 
Motivated by these advances, Razavian et al. \citep{razavian_visual_2015} conducted extensive experiments on instance retrieval datasets using CNN features and produced promising results. The \textit{regional maximum activations of convolutions} (R-MAC) \cite{tolias_particular_2016} feature vector is constructed by encoding the image regions into a compact supervector with scale and translation invariances. Subsequently, the region descriptors are aggregated (sum) and L2-normalized. The representation, when paired with re-ranking and query expansion, matched the state-of-the-art in retrieval accuracy. R-MAC became a prominent method and inspired several extensions. Along a similar line of investigation, Wang et. al. \cite{wang_aggregating_2023} proposed a compact representation named multi-models deep feature aggregation (MDFA) which effectively eliminated background noise and improved image retrieval accuracy. 

\textbf{Fisher-encoded CNN Descriptors:}
Fisher Vectors (FV) have a significant advantage over CNNs in that they are relatively less expensive to train and evaluate. Uricchio et. al. \citep{uricchio_fisher_2015} computed CNN output activations from few hundreds of proposal windows to generate a multi-scale final representation using the Fisher Vector encoding and Principal Component Analysis (PCA). Perronnin \cite{perronnin_fisher_2015} suggested a hybrid architecture for image classification that incorporates the best features of Fisher and CNN pipelines. The first layers of the architecture are unsupervised and involve Fisher Vector computation and dimension reduction. The successive layers are supervised, fully connected and trained using back-propagation. Typically, Fisher Vector is computed using non-binary SIFT features. Examples of accurate retrieval and sub-optimal retrieval outcomes by baseline Fisher Vector (FV) on INRIA Holidays dataset are shown in Fig. \ref{fv-accurate} and \ref{fv-suboptimal}. Amato et al. \citep{amato_combining_2016} combined CNN with binary features and termed the proposed representation BMM-FV as it used a Bernoulli Mixture Model (BMM) for describing binary outcomes. Additionally, Fisher Vectors have been recommended for Siamese neural networks and used for aggregating CNN image descriptors \cite{ong_siamese_2017}. The representation produced significant improvements in retrieval scores and was robust and discriminative. Although a deep representation is quite effective, it requires much computing power for the optimization and a lot of training data for generalization.
\begin{figure*}[htp]
	\centering
	\includegraphics[width=0.9\textwidth]{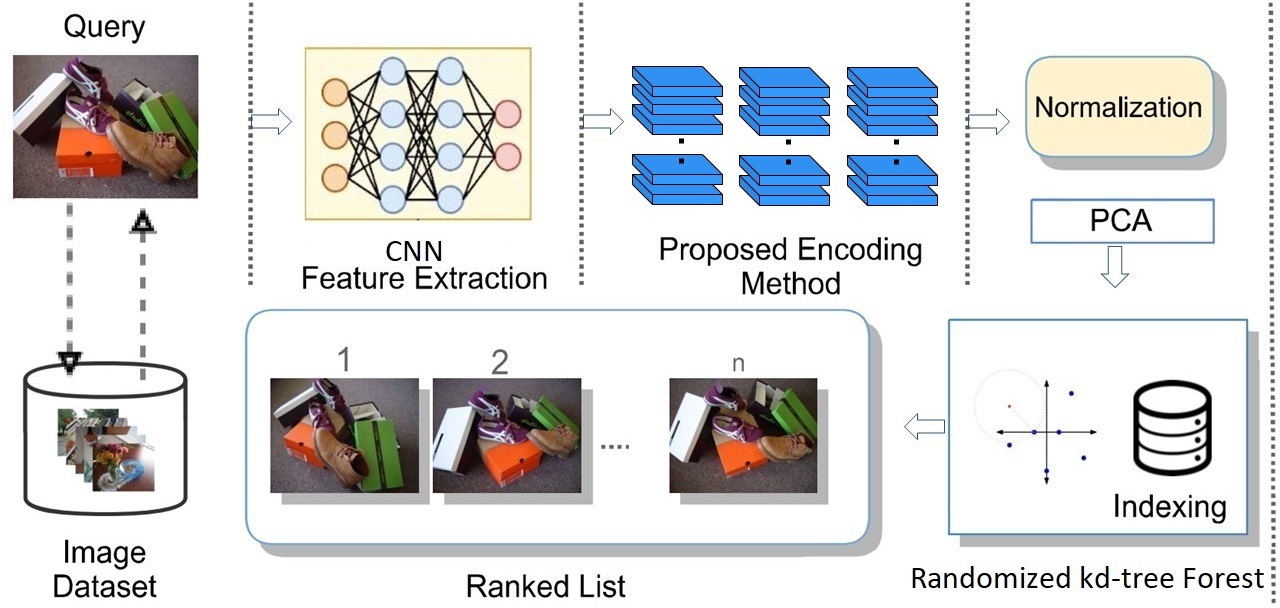}
	\caption[The scalable retrieval framework using the proposed vector representation for retrieval performance evaluation.]{The scalable image retrieval framework using the proposed vector representation for retrieval performance evaluation.}
	\label{fv-cnn-pipeline}
\end{figure*}
\section{Proposed Method}
\label{sec3}
In this section we introduce our multi-scale aggregation and pooling scheme as well as the scalable retrieval framework. 

ConvNets are considered state-of-the-art and have established themselves as highly successful classifiers for vision tasks. Their capacity to learn highly discriminative activation (feature) maps from image regions is what contributes mostly to their success. Formally, if $x_{ij}$ is the feature vector at location $(i; j)$ in a specific layer, and $y_{ij}$ for the following layer, a non-linear activation layer computes the feature maps $y_{ij}^{map}$ as: 

\begin{equation}
y^{map} = f_{ks}({\{x_{{si} + \delta i, sj + \delta j}\}_{0 \leqslant \delta i, \delta j \leqslant kernel}}),
\label{eqnconvnet}
\end{equation}

where \textit{kernel}, \textit{s}, and $f_{ks}$ stand for ``kernel size", ``stride", and ``layer type" respectively. As layers stacked together, \textit{convolution} or \textit{average pooling} essentially conducts element-wise matrix multiplication, \textit{max pooling} downsamples the input using the spatial Max() operation, \textit{Rectified Linear Unit} (ReLU) introduces non-linearity to the model, and so forth. 

The effectiveness of global CNN activations in classification and retrieval scenarios is contrained by a lack of geometric invariance. We demonstrate the invariance of CNN activations using multi-scale activations and aggregate the convolutional features in a second-order fashion using a Fisher Kernel framework \citep{perronnin_improving_2010}. Popular representations such as Spatial Pyramid Matching \citep{lazebnik_beyond_2006} are ordered representations of the scene but work only at a single scale. As an alternative, we extract activations from three different resolutions of the input image to improve their invariance without compromising their discriminative efficacy. We employ the pre-trained CNN VGG-16 \footnote{https://www.vlfeat.org/matconvnet/} as it has been shown to enhance the derived scene representation in well-established scene retrieval datasets while incurring lower computational demand \citep{tolias_particular_2016}. After the ReLU transformation, we utilize the 4096-dimensional activations of the final fully connected layer of the VGG-16 \cite{simonyan_very_2015}. Additionally, we experimented with a deeper architecture from the VGG family, the VGG-19 network. VGG networks are known to suffer from vanishing gradients as their depth increases. Residual Networks (ResNets) \cite{he_deep_2016} address this problem by introducing residual connections, which facilitate better gradient flow through deeper networks with reduced complexity. To ensure a fair comparison, we use the 512-dimensional activations from the final fully connected layer of two ResNet versions - ResNet-18, a shallower model, and ResNet-34, a deeper model. All four models were pretrained on ImageNet. 

More specifically, let $ \mathcal{X} = \{x_1, x_2,...x_n\} $, $x_j$ $\in \mathbbm{R^{\textit{d}}} $ denote the descriptors (\textit{e.g.} a set of Dense SIFT \cite{lowe_distinctive_2004}, or CNN \cite{simonyan_very_2015} descriptors) extracted from an image. A Gaussian Mixture Model (GMM) learned with Expectation-Maximization (EM) models the feature distribution. Let $ \theta = \{(\mu_{i}^{\theta}, \sum_{i}, \pi_{i}), i=1,2, ... k\} $ be a GMM (with $\textit{k}$ Gaussian modes), where $\mu_i^{\theta}$, $\sum_i$ and $\pi_i$ are, respectively, the \textit{means}, \textit{covariance matrices}, and \textit{prior probabilities} of the \textit{i}-th mode computed on the (independent) training set. The \textit{posterior probability} that $x_j$ is associated to a mode $\textit{i}$ is given by the soft-assignment:

\begin{equation}
q_{ji} = \frac{\exp [-\frac {1}{2}(x_j - \mu_{i}^\theta)^T \sum_{k}^{-1}(x_j - \mu_{i}^\theta)]}{\sum_{t=1}^{k} \exp [-\frac {1}{2}(x_j - \mu_t^\theta)^T \sum_{k}^{-1}(x_j - \mu_t^\theta)]}.  
\end{equation}

We compute the \textit{mean} and diagonal \textit{covariance} vectors for each mode $\textit{k}$ as: 

\begin{equation}
u_{jk} = \frac{1}{N \sqrt{\pi_{k}}} \sum_{i=1}^N q_{ik} \hspace{1mm} \frac{x_{ji} - \mu_{jk}^\theta}{\sigma_{jk}},
\end{equation}

\begin{equation}
v_{jk} = \frac{1}{N \sqrt{2\pi_{k}}} \sum_{i=1}^N  q_{ik} \bigg[{\Big(\frac{x_{ji} - \mu_{jk}^\theta}{\sigma_{jk}}\Big)}^2 - 1\bigg],
\end{equation}

where \textit{j} = \{1, 2, 3, ... D\}. Finally, the \textit{means} ${u_{k}}$ ($1^{st}$-order statistic) and the diagonal \textit{covariance matrices} ${v_{k}}$ ($2^{nd}$-order statistic) are concatenated into a supervector:
\begin{equation}
\phi=[u_1^T, u_2^T, ...u_{k}^T, v_1^T, v_2^T, ... v_{k}^T]^T, \qquad u_{i}, v_{i}  \in \mathbbm{R^{{d}}}.
\label{eqnV}
\end{equation}



The encoded vector \textit{v} is then power-normalized and $L2$-normalized as last steps. \textit{Power-normalization} enhances the representation's discriminative power by diminishing the impact of bursty elements brought on by repetitive structures \cite{jegou_aggregating_2012}. Each component $v_j$ is changed into:

\begin{equation}
\phi_j  = sign(\phi_{j}) \times |\phi_{j}|^{\alpha}, \qquad j \in [1,2, ..., D],
\end{equation}

where $0$ $\leq$ $\alpha$ $<1$ ($\alpha$ is a normalization parameter). This step needs tweaking for best performance (near-optimal options being $\alpha$ = 0.2, 0.4 or 0.5). \textit{L2-normalization} makes the representation invariant to the number of features. The updated vector $\phi$ is divided by its L2-norm: $\phi = \frac{\phi}{\norm{\phi}_{2}}$.



The \textit{full-dimensional} proposed representation for the descriptor set $ \mathcal{X} $ at one scale is given by $\phi$. To produce the final representation, the representations from the multiple scales are average pooled. The steps of the proposed multi-scale vector encoding and pooling algorithm are depicted in Fig. \ref{fv-cnn-encoding}.  To show that the proposed encoding scheme is effective, we run visual search experiments on benchmarks and assess its performance. We experiment with dimension reduction as well, and PCA is used to analyze performance at reduced dimensions. A scalable scene retrieval framework that was developed to employ the proposed representation is shown in Fig. \ref{fv-cnn-pipeline}. The framework leverages \textit{randomized kd-tree forest} \footnote{https://www.vlfeat.org/overview/kdtree.html} for rapid approximate nearest neighbor searches in high-dimensions and efficient indexing to return ranked results.

\section{Experimental Results and Discussion}
\label{sec4}

In this section, we introduce the two benchmark datasets and outline the evaluation metrics to assess the effectiveness of our method. We then describe the ablation studies conducted to investigate the contributions of different components. Additionally, we compare our method to state-of-the-art techniques, analyze the results, and provide an in-depth discussion of the performance.


\subsection{Evaluation Benchmark and Metrics}
\label{chap-4datasets}

We set up our scene retrieval experiment in accordance with the baseline technique experimental setup \citep{perronnin_improving_2010} and \citep{perronnin_fisher_2015} (since they proposed similar compact representations for retrieval and classification tasks). The two retrieval benchmarks that serve as the basis for our method's evaluation are standard and publicly available. 



\subsubsection{INRIA Holidays}
The INRIA Holidays dataset \footnote{https://lear.inrialpes.fr/$\sim$jegou/data.php}
\cite{jegou_improving_2010} is made up of 1491 photos that have been split into 500 groups, each of which contains a unique scene or a collection of objects. The dataset consists of 500 queries and 4.456 M descriptors. The results are presented as Mean Average Precision (mAP) over the 500 queries, with the query omitted from the ranking list (leave-one-out approach). 

\subsubsection{University of Kentucky Benchmark (UKBench)}
The UKBench \footnote{https://archive.org/details/ukbench} \cite{wengert_bag--colors_2011} has 10,200 images of 640x480 resolution that relate to 2,550 different objects and scenes of diverse categories. Performance is assessed using the N-S score (range: 0-4), the average number of images returned in the first 4 places (4$\times$Recall@4). 


\subsection{Impact of Parameters}
\label{chap4-impact}
\begin{figure*}[htp!]
	\centering
	\includegraphics[width=0.85\textwidth]{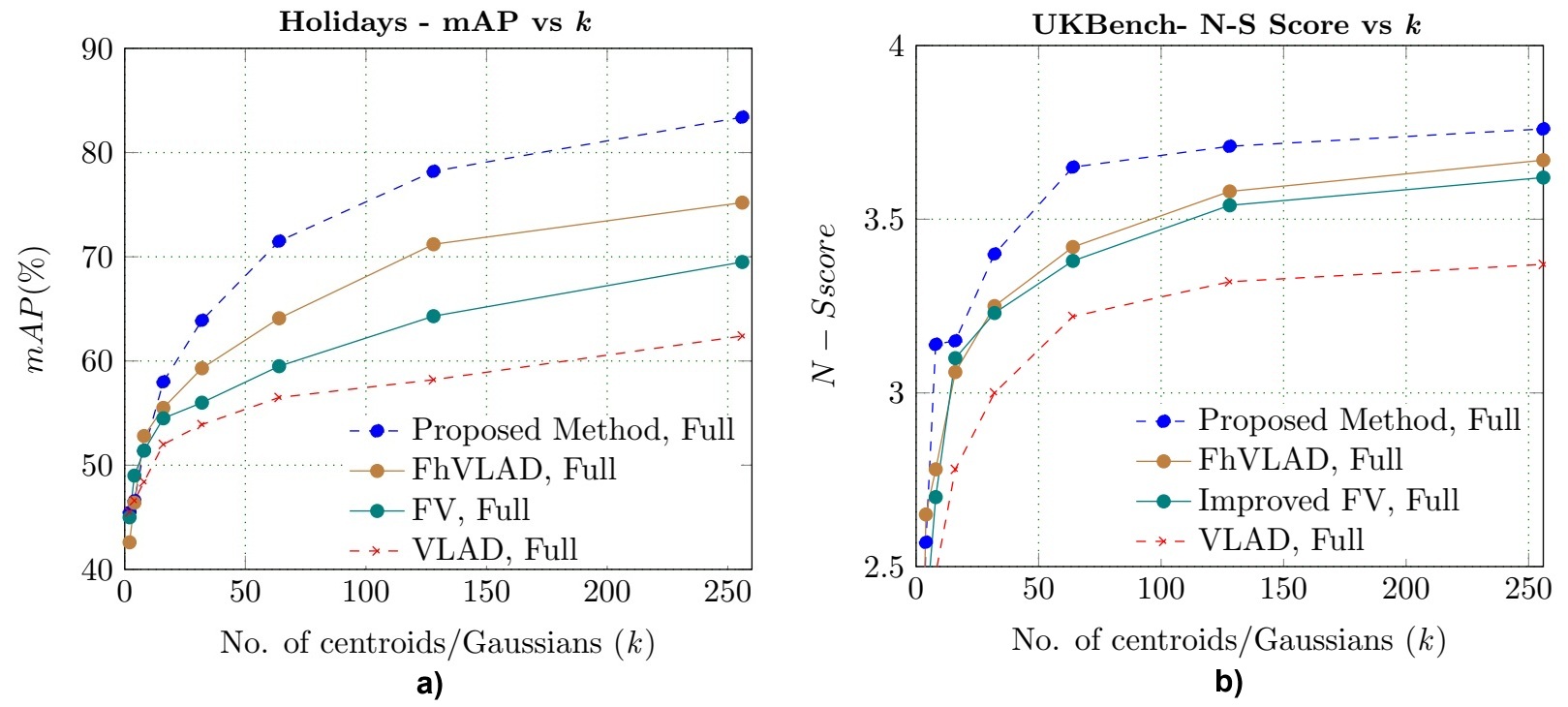}
	\caption[Performance comparison with baselines: Impact of number of Gaussian components \textit{k} a) on Mean Average Precision(mAP) on Holidays, b) on N-S Score on UKBench.]{Performance comparison with baselines. Impact of number of Gaussian components \textit{k} a) on Mean Average Precision(mAP) on Holidays, b) on N-S Score on UKBench. (Higher score is preferable). Parameter $\alpha = 0.5$. }
	\label{mapbaseline}
\end{figure*}

\textbf{Impact of number of Gaussian Components}: In order to determine how the retrieval performance varies with disparate Gaussian components ($\textit{k}$), we examine the performance of our method with varying component sizes and compare it to our baselines VLAD \citep{jegou_aggregating_2012} and FV \citep{perronnin_improving_2010}. Fig. \ref{mapbaseline} shows the impact of \textit{k} on the scores on Holidays and UKBench for the full-sized proposed descriptor. It is noticed that, up until a certain point, the retrieval performance grows as the number of Gaussians increases. For \textit{k} = 128, we achieve a 74.5\% mAP on Holidays, and an 3.40 N-S score on UKBench. For larger values of \textit{k}, however, the computational cost becomes impractical and the dimensions becomes too large for efficient indexing and retrieval. Therefore, to obtain a reasonable performance-efficiency trade-off, we limit the size of \textit{k} to 128 and restrict our analysis to \textit{k}=256. 

\begin{figure*}[htp!]
	\centering
	\includegraphics[width=0.85\textwidth]{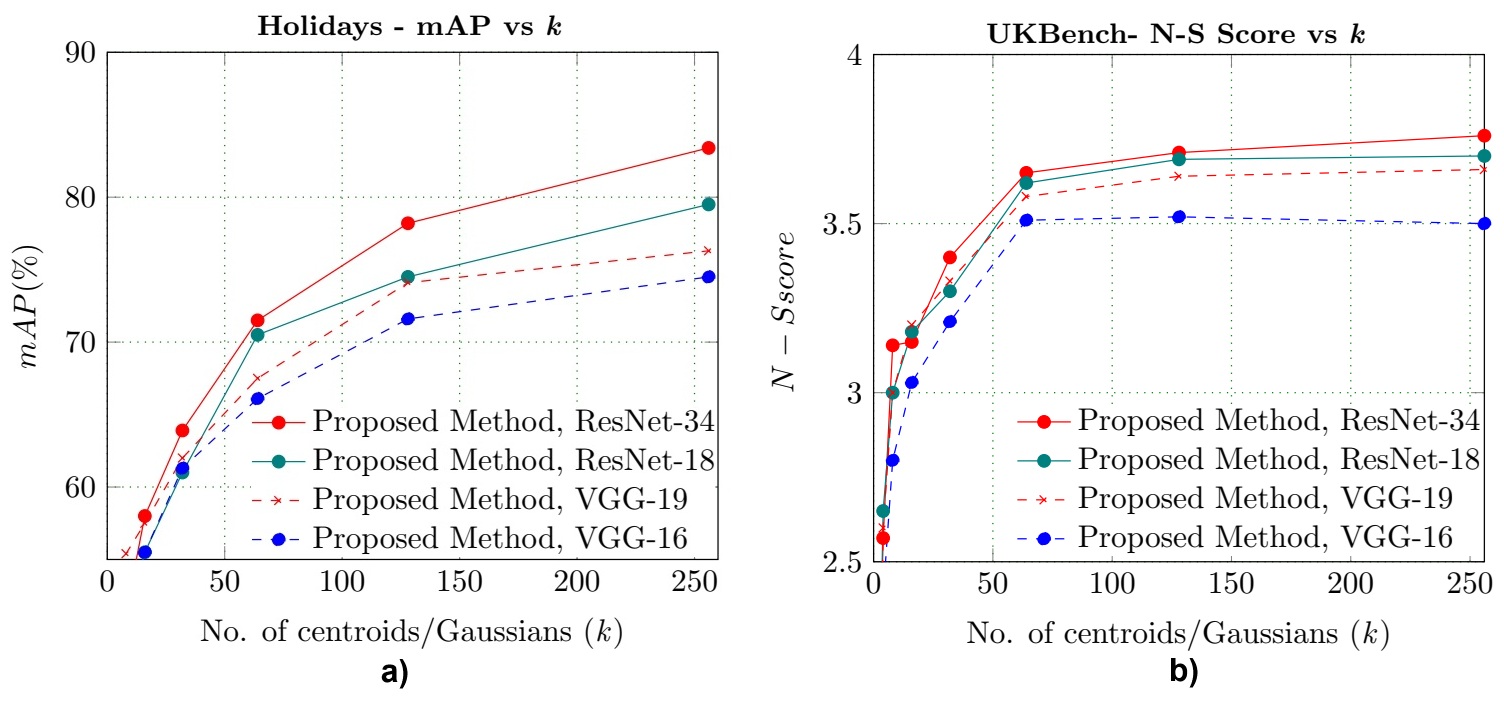}
	\caption[Ablation experiments using four different deep learning architectures: Impact of number of Gaussian components \textit{k} a) on Mean Average Precision(mAP) on Holidays, b) on N-S Score on UKBench.]{Ablation experiments using four different deep learning architectures: Impact of number of Gaussian components \textit{k} a) on Mean Average Precision(mAP) on Holidays, b) on N-S Score on UKBench. (Higher score is preferable). Parameter $\alpha = 0.5$. }
	\label{ablation-fig}
\end{figure*}


\subsection{Ablation Experiments}
\label{chap4-ablation}

We conducted ablation experiments using four different deep learning architectures, as described in Section \ref{sec3}. Table \ref{table-ablation} summarizes the result of these experiments, where feature maps were extracted from the various models. We tested our method for different values of Gaussian components (\textit{k}). As shown in Table \ref{table-ablation}, the best performance and robustness in retrieval accuracy is achieved using a dimension size of 256 with the ResNet-34 architecture. Additionally, our experiments also demonstrate that deeper networks improve the image retrieval accuracy. Specifically, the ResNet-34 architecture outperforms the others, making it the most suitable model among the four pre-trained models evaluated. Notably, the VGG networks consistently improve upon the baseline FV \cite{perronnin_fisher_2007} and FhVLAD \cite{bhowmick_fhvlad_2021} scores. It can be inferred that the aggregation of second-order information of multi-scale CNN features within the Fisher Kernel framework leads to significant gains (as seen in Figure \ref{ablation-fig}). The models are pre-trained on ImageNet dataset. Our experiments were done on a platform with AMD Ryzen 9 CPU (4.30 GHz, 8 cores), 32 GB of RAM, and an NVIDIA GeForce RTX 2060 GPU with 6 GB of memory. \\

\begin{table}[tp]

  \renewcommand{\arraystretch}{1.3}
  \caption{Ablation experiments comparing results when using different Deep Learning architectures. \textit{k} refers to the number of Gaussian components.}
  \label{table-ablation}
  \centering
  \begin{tabular}{@{}p{25mm}>{$}c<{$}@{\hskip1.5em}S[table-format=1.1]S[table-format=1.1]@{}}
    \toprule
    \multicolumn{2}{@{}l}{\bfseries\makecell{CNN Architecture}} & {\thead{\textit{Holidays} \\ (mAP(\%))}} & {\thead{\textit{UKB} \\ (Score/4)}} \\
    \otoprule
    \multirowcell{4}{VGG-16} & k=32 & 61.3 & 3.21 \\
                             & k=64 & 66.1 & 3.51 \\
                             & k=128 & 71.6 & 3.52 \\
                             & k=256 & \textbf{74.5} & \textbf{3.50} \\
    \midrule
    \multirowcell{4}{VGG-19} & k=32 & 62.0 & 3.33 \\
                             & k=64 & 67.5 & 3.58 \\
                             & k=128 & 74.1 & 3.64 \\
                             & k=256 & \textbf{76.3} & \textbf{3.66} \\
    \midrule
    \multirowcell{4}{ResNet-18} & k=32 & 61.0 & 3.30 \\
                                & k=64 & 70.5 & 3.62 \\
                                & k=128 & 74.5 & 3.69 \\
                                & k=256 & \textbf{79.5} & \textbf{3.70} \\
    \midrule
    \multirowcell{4}{ResNet-34} & k=32 & 63.9 & 3.4 \\
                                & k=64 & 71.5 & 3.65 \\
                                & k=128 & 78.2 & 3.71 \\
                                & k=256 & \textbf{83.4} & \textbf{3.76} \\
    \bottomrule
  \end{tabular}
\end{table}

\subsection{Comparison with State-of-the-art}
\label{chap4-comparison}


\begin{table*}[!htbp]
\caption[Performance comparison of the full-size proposed supervector with state-of-the-art on Holidays and UKBench.]{Performance comparison of the full-size proposed supervector with state-of-the-art on Holidays and UKBench. \vspace{2mm} 
	\centering
 
	\begin{tabular}{p{4.3cm} p{1.7cm} p{1.1cm}p{1.7cm}p{1.7cm}} 
		\toprule
		Method & Codebook \newline size & D & \textit{Holidays} \hspace{1mm} (mAP(\%)) & \textit{UKBench} \hspace{1mm} (score/4) \\
		\midrule \addlinespace
		Improved FV \cite{perronnin_large-scale_2010} & \textit{k}=64 & 16384 & 59.5 & 3.25 \\ 
		VLAD+SSR \cite{jegou_aggregating_2012} & \textit{k}=64 & 16384 & 59.8 & 3.28 \\ 
		VLAD + Intra + Adapt \cite{arandjelovic_all_2013} & \textit{k}=64 & 8192 & 64.6 & - \\ 
		VLAD* \cite{delhumeau_revisiting_2013} & \textit{k}=64 & 8192 & 62.2 & - \\ 
		HVLAD \cite{liu_making_2016} & \textit{k}=64 & 8192 & 66.8 & 3.48 \\ 
		Triangulation Embedding \cite{jegou_triangulation_2014} & \textit{k}=64 & 8192 & 75.6 & - \\ 
		CEVLAD \cite{zhou_distribution_2016} & \textit{k}=64 & 8192 & 67.6 & 3.43 \\ 
		Fine Residual VLAD \cite{liu_fine-residual_2016} & \textit{k}=64 & 8192 & 61.4 & 3.38 \\
		RVD-W \cite{husain_improving_2017} & - & 512 & 69.0 & - \\
		SDEVLAD \citep{liu_adding_2018} & \textit{k}=64 & 8256 & 71.4 & - \\ 
        FhVLAD \citep{bhowmick_fhvlad_2021} & \textit{k}=64 & 8256 & 67.9 & 3.51 \\
		OxfordNet \citep{ng_exploiting_2015} & - & 512 & 83.8 & - \\
		NetVLAD \citep{arandjelovic_netvlad_2018} & \textit{k}=64 & 512 & 79.9 & - \\
        MDFA3 \citep{wang_aggregating_2023} & \textit{k}=64 & 512 & 87.15 & - \\
		\midrule
		Our proposed method & - & 16384 & \textbf{83.42} & \textbf{3.76} \\

		\bottomrule \addlinespace
	\end{tabular} 
	
	} 
	\label{table:res1-chap4-cnn}
\end{table*}

\begin{table*}[!htbp]
	\centering
 \caption{Performance comparison of the PCA-reduced (D\textquotesingle = 128\hspace{0.5mm}) proposed supervector on Holidays and UKBench with state-of-the-art.} 

		\begin{tabular}{p{4.5cm} p{1.0cm}p{1.0cm}p{1.7cm}p{1.7cm}} \toprule
		Method $\downarrow$ & k & D' & Holidays & UKBench \\ 
		\midrule 
		
		Improved FV \citep{perronnin_large-scale_2010} 
		& 64 & 128 & 56.5 & 3.33 \\ 

		
		VLAD + SSR \citep{jegou_aggregating_2012}  
		& 64 & 128 & 55.7 & 3.35 \\  
		
		Multivoc-VLAD \citep{jegou_negative_2012}  
		& 64 & 128 & 61.4 & 3.36 \\  
		
		VLAD+intra+adapt \citep{arandjelovic_all_2013}  
		& 64 & 128 & 62.5 & - \\  
		
		HVLAD \citep{liu_making_2016}  
		& 256 & 128 & 64.0 & 3.4 \\  
		
		Temb \citep{jegou_triangulation_2014}  
		& 256 & 128 & 65.7 & - \\  
		
		RVD-W \citep{husain_improving_2017} 
		& - & 128 & 61.6 & - \\  
		
		Fine-residual VLAD \citep{liu_fine-residual_2016}  
		& 256 & 128 & 62.2 & 3.43 \\  
		
		SDEVLAD \citep{liu_adding_2018}  
		& 256 & 128 & 73.6 & - \\  
		
		
		OxfordNet \citep{ng_exploiting_2015}  
		& - & 128 & 81.6 & - \\  
		
		
		\hline 
		
		Our proposed method 
		& - & 128 & \textbf{74.61} & \textbf{3.48} \\
		\bottomrule
	\end{tabular}
	\label{table:res2-chap4-cnn}
\end{table*}


For an even comparison Table \ref{table:res1-chap4-cnn} compares our full-size descriptor's performance (with ResNet-34 model) with that of more recent, advanced encoding methods. It is noted that the proposed method obtains 83.42\% mAP on Holidays and 3.76 N-S score on UKBench. This raises the performance over the recent baseline FhVLAD \cite{bhowmick_fhvlad_2021} by 15.5 \% on Holidays and 0.25 increase in N-S score on UKBench. Our method still lags behind recent techniques such as the MDFA3 \citep{wang_aggregating_2023} in retrieval accuracy. MDFA3 extracts deep features from two pre-trained CNN models, which are then aggregated into a robust supervector inspired by the attention mechanism. Triangulation Embedding (Temb) \citep{jegou_triangulation_2014} which is based on triangulation strategy for embedding and democratic aggregation of descriptors has results that are at par with our result on PCA-reduced supervectors on the Holidays dataset (Please see Table \ref{table:res2-chap4-cnn}). We believe the impact of the discriminative power of convolutional features and the effective second-order encoding strategy is what causes the performance boost in the proposed descriptor. 

\begin{figure*}[htp]
	\centering
	\includegraphics[width=0.92\textwidth]{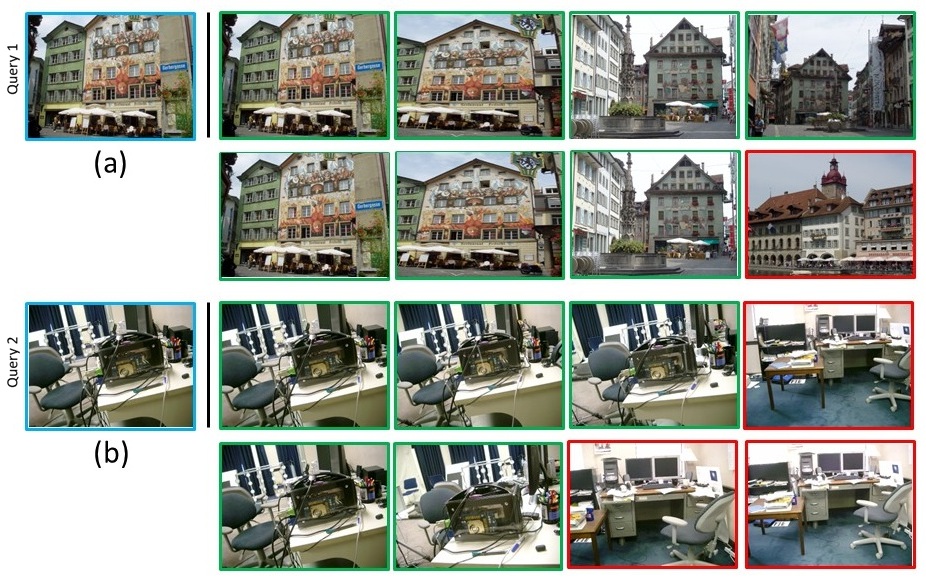}
	\caption[Example retrieval results from the Holidays dataset and the UKB dataset: For query (a) in the blue box, the top row corresponds to our results on Holidays, bottom row corresponds to results obtained by recent FhVLAD \cite{bhowmick_fhvlad_2021}. For query (b) also in the blue box, the top row corresponds to our results on the UKB dataset, and the bottom row corresponds to baseline results obtained by the recent FhVLAD \cite{bhowmick_fhvlad_2021}. A blue box \fcolorbox{teal}{white}{\phantom{\rule{5pt}{5pt}}} surrounds each query. A green box \fcolorbox{green}{white}{\phantom{\rule{5pt}{5pt}}} corresponds to results that match with the ground-truth. Results that do not reflect groundtruth (or are edge cases) are in a red box \fcolorbox{red}{white}{\phantom{\rule{5pt}{5pt}}}.]{Example retrieval results from the Holidays dataset and the UKB dataset: For query (a) in the blue box, the top row corresponds to our results on Holidays, bottom row corresponds to baseline results from recent FhVLAD \cite{bhowmick_fhvlad_2021}. For query (b) also in the blue box, the top row corresponds to our results on the UKB dataset, and the bottom row corresponds to baseline results from the recent FhVLAD \cite{bhowmick_fhvlad_2021}. In (b), a failure instance of our approach is evident, as the fourth instance in the top row does not match the ground truth. A blue box \fcolorbox{teal}{white}{\phantom{\rule{5pt}{5pt}}} surrounds each query. A green box \fcolorbox{green}{white}{\phantom{\rule{5pt}{5pt}}} corresponds to results that match with the ground-truth. Results that do not reflect groundtruth (or are edge cases) are in a red box \fcolorbox{red}{white}{\phantom{\rule{5pt}{5pt}}}.}
	\label{Hol-UKB-results}
\end{figure*}

The proposed descriptor demonstrates consistent superior performance against baselines FV \citep{perronnin_improving_2010}, SDEVLAD \citep{liu_adding_2018}, and FhVLAD \cite{bhowmick_fhvlad_2021}, as well as several recent SIFT and CNN-based extensions (as seen in Table \ref{table:res1-chap4-cnn}). NetVLAD's \citep{arandjelovic_netvlad_2018} high accuracy is mainly attributed to replicating VLAD in a CNN framework and the trainable generalized VLAD layer providing high distinctiveness. Compared to sophisticated CNN-based representations like NetVLAD \citep{arandjelovic_netvlad_2018} and OxfordNet \citep{ng_exploiting_2015}, the proposed representation vector's performance is noteworthy because \textit{i)} The method is codebook-free and relies on rudimentary vector operations of pre-trained convolutional features, and \textit{ii)} It only requires a few parameters, with the most important ones being - number of Gaussian components \textit{k} and $\alpha$. For both the datasets under consideration, the proposed descriptor has a running time $\mathcal{O}(k(2d+1))$ and delivers a competitive accuracy at decreased dimensions (Table \ref{table:res2-chap4-cnn}). Two query results of image retrieval are shown in Fig. \ref{Hol-UKB-results}, one from Holidays dataset and other from UKB. Both queries involve challenge related to illumination variation,  viewpoint  variance, and clutter. While our method shows improved robustness in such scenarios, occasional inaccuracies persist, highlighting areas for future enhancement as discussed in the following section.


\section{Limitations and Future Work}
\label{sec5}

Notwithstanding the positive results of the paper, there are still some challenges and limitations that need to be discussed. We chose the Fisher Kernel due to its hallmark characteristics and merits. However, it does have demerits that our approach inherits, such as hyper parameter tuning and increased computational cost of encoding, which can be challenging and time-consuming. The computational complexity can be pronounced when we consider larger number of Gaussian components (e.g. \textit{k} = 1024). In high dimensionality the approach is also susceptible to the curse of dimensionality. The use of multi-scale features may also make it incompatible with techniques that require spatial verification, although this remains to be further explored.

To address these limitations we recommend exploring advanced aggregation mechanisms and also exploring those ideas to design vector aggregation methods for other domains such as medical image classification and data compression. In the future, determining the best architecture for retrieval tasks can be done at a broader level, considering the wide range of architectures that are available. A key direction lies in revisiting deeper architectures, especially for large-scale retrieval tasks and other complex tasks. While one may aim to get better performance with deeper architectures like ResNet-101, however their computational requirements become significantly higher as they are too deep. This also results in slower feature extraction process and a steep increase in resource consumption. Therefore, it is crucial to look for a trade-off between depth and efficiency to optimize performance. As part of our future work, we plan to explore the applicability of the proposed method in road scene object classification, as well as extend its use to other domains such as the retrieval of cultural heritage art and paintings.

\section{Conclusion}
\label{sec6}
We described the design of a novel compact supervector for scene retrieval in this paper. The proposed method is built on the Fisher Kernel framework for effective second order encoding of the multi-scale CNN features to enhance the representational capacity. The limited increase in dimension and computational complexity caused by the discriminative encoding is a trade-off. Experimental evaluation on two well-known datasets show our method yields significant gain over recent CNN-based and traditional SIFT-based signatures in retrieval performance. The proposed vector is also expected to improve image and object classification performance.


\begin{acknowledgements} The authors express their gratitude to Dr. Sarbani Palit, CVPR Unit, ISI Kolkata for her valuable suggestions and comments at the final stage of this work.  
\end{acknowledgements}

\section{Statements and Declarations}
 \textbf{Ethical Approval}\\
This declaration is not applicable. \\ \noindent
\textbf{Competing interests} \\
The authors have no competing interests to declare that are relevant to the content of this article. \\ \noindent
\textbf{Authors' contributions} \\
All authors contributed to the study conception and design. Author 1 investigated the presented idea, performed the experiments, designed the figures and drafted the main manuscript. 
Author 2 revised the manuscript versions critically and contributed in interpretation of data.
Author 3 provided support in drafting the manuscript, revised the manuscript versions critically, contributed in interpretation of data, and supervised the work. All authors read and approved the final manuscript. \\ \noindent
\textbf{Funding} \\
The authors did not receive support from any organization for the submitted work. \\ \noindent
\textbf{Availability of data and materials} \\
The datasets generated analysed during the current study are available in the following repositories: \\ \href{https://lear.inrialpes.fr/~jegou/data.php}{https://lear.inrialpes.fr/~jegou/data.php}, \\
\href{https://archive.org/details/ukbench}{https://archive.org/details/ukbench}

%
\bibliographystyle{spbasic}      

\bibliography{SNCS-2025.bib}    

\end{document}